\documentclass[10pt,twocolumn,letterpaper]{article}
\usepackage{iccv}
\usepackage{times}
\usepackage{epsfig}
\usepackage{graphicx}
\usepackage{amsmath}
\usepackage{amssymb}

\usepackage{adjustbox}
\usepackage{pifont}
\usepackage{caption}
\usepackage{subcaption}
\usepackage{dsfont}
\usepackage{multirow}
\usepackage{tablefootnote}
\usepackage{arydshln}

\usepackage[pagebackref=true,breaklinks=true,letterpaper=true,colorlinks,bookmarks=false]{hyperref}

\iccvfinalcopy 
\def\iccvPaperID{6848} 

\ificcvfinal\fi

\begin{document}

\title{Aug3D-RPN: Improving Monocular 3D Object Detection \\
by Synthetic Images with Virtual Depth}

\author{Chenhang He$^{1,2}$,  Jianqiang Huang$^{2}$,  Xian-Sheng Hua$^{2}$,  Lei Zhang$^{1,2}$
\thanks{Corresponding author. }\\
{\tt\small \{csche\}@comp.polyu.edu.hk, jianqiang.hjq@alibaba-inc.com}\\
{\tt\small huaxiansheng@gmail.com, cslzhang@comp.polyu.edu.hk}
}

\maketitle

\begin{abstract}
Current geometry-based monocular 3D object detection models can efficiently detect objects by leveraging perspective geometry, but their performance is limited due to the absence of accurate depth information.
Though this issue can be alleviated in a depth-based model where a depth estimation module is plugged to predict depth information before 3D box reasoning, the introduction of such module dramatically reduces the detection speed. Instead of training a costly depth estimator, we propose a rendering module to augment the training data by synthesizing images with virtual-depths. The rendering module takes as input the RGB image and its corresponding sparse depth image, outputs a variety of photo-realistic synthetic images, from which the detection model can learn more discriminative features to adapt to the depth changes of the objects. Besides, we introduce an auxiliary module to improve the detection model by jointly optimizing it through a depth estimation task. Both modules are working in the training time and no extra computation will be introduced to the detection model. Experiments show that by working with our proposed modules, a geometry-based model can represent the leading accuracy on the KITTI 3D detection benchmark. 
\end{abstract}

\section{Introduction}
\label{sec:intro}
3D object detection is one of the key technologies in autonomous driving. Detecting and localizing the objects accurately in the 3D space is critical for the autonomous vehicles to run safely over a long distance. While the use of high-end LiDAR for high-precision 3D object localization \cite{voxelnet, second, pointrcnn, pointpillars, std, sassd, pvrcnn} has drawn much interest in academia and industry, the demand for more economic monocular based alternatives is also increasing, and many monocular 3D object detection methods
\cite{m3d-rpn, smoke, geo3d, am3d, monet3d, monogrnet, monofenet, monopair, monopsr, d4lcn,patchnet, pseudo-e2e, pseudo, pseudo++, oftnet, mlf} have been developed in the past years. 

\begin{figure}[t]

\begin{subfigure}{0.49\columnwidth}
\centering
\includegraphics[width=\columnwidth]{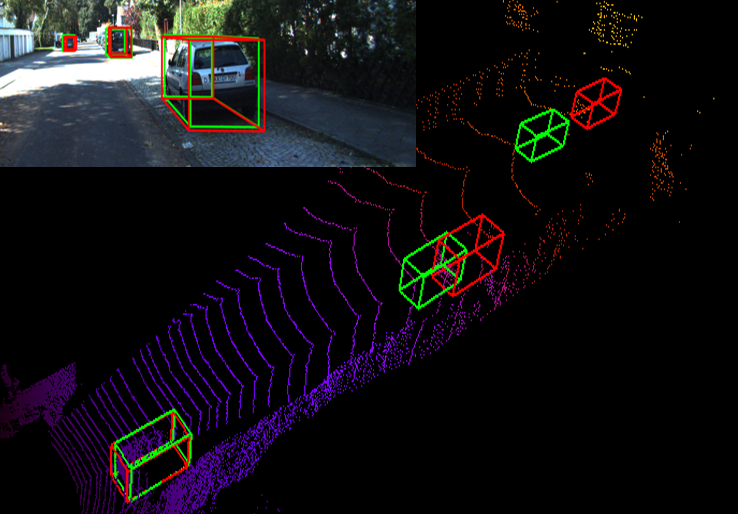}
\vspace{-16pt}
\caption{}
\end{subfigure}
\begin{subfigure}{0.49\columnwidth}
\centering
\includegraphics[width=\columnwidth]{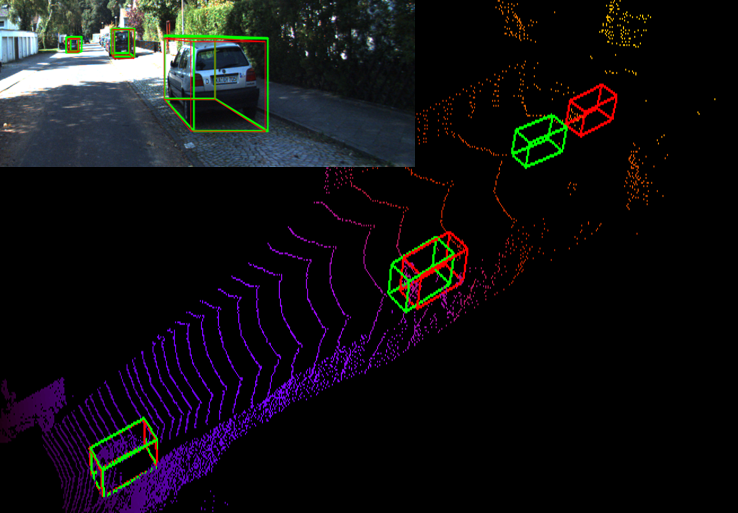}
\vspace{-16pt}
\caption{}
\end{subfigure}

\caption{(a) Bounding box predictions by \textit{M3D-RPN} \cite{m3d-rpn}, where the model can accurately predict the size and orientation of the objects, but fail to accurately localize the objects at far distances. (b) Bounding box predictions by \textit{M3D-RPN}, learnt with synthetic virtual-depth images. The predicted and ground-truth boxes are shown in green and red, respectively.}
\label{fig:concept}
\vspace{-4mm}
\end{figure}

Current researches on monocular 3D object detection can be categorized into two streams, i.e., geometry-based methods and depth-based methods. Geometry-based methods \cite{m3d-rpn, geo3d, shift-rcnn, smoke, monogrnet, monet3d, monopair, rtm3d} mainly focus on predicting various geometric primitives based on perspective transformation and detecting the objects to fit these primitives. Good geometric primitives can establish a robust connection between the 3D object predictions and 2D image features. Figure \ref{fig:concept}(a) shows an example, where the geometric relationship between 2D and 3D anchors \cite{m3d-rpn} is used to learn the object size and orientation from image semantics. However, such methods are difficult to estimate the object depth from a single image. As a result, the geometry-based methods are not able to accurately localize the objects at a long distance. 

\begin{figure*}[t]
\centering
\includegraphics[width=1\textwidth]{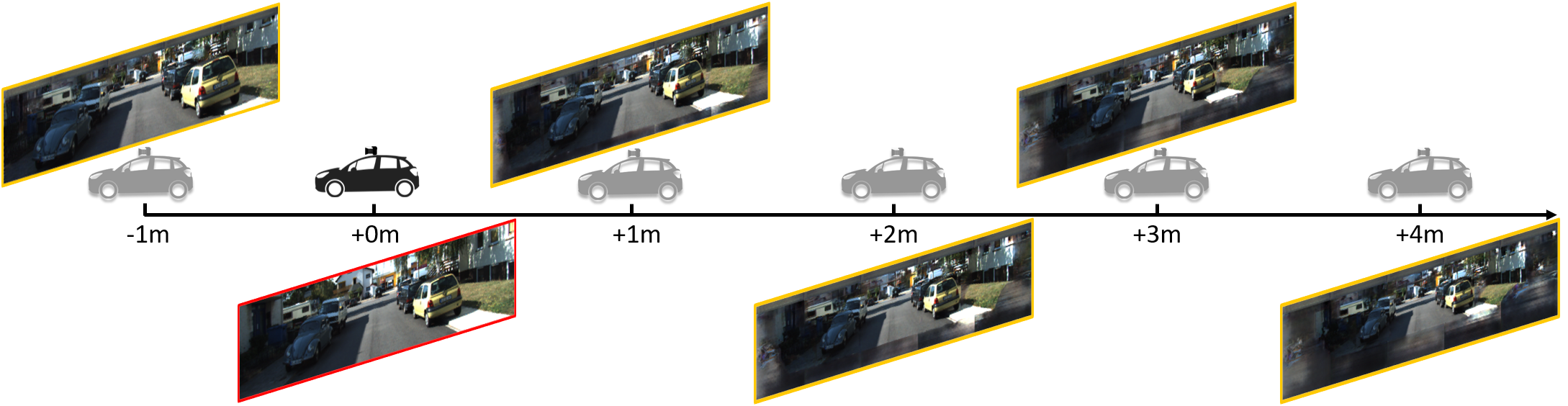}
\caption{Examples of reference image (in red) and its corresponding synthetic virtual-depth images (in yellow) with relative virtual depth -1, +1, +2, +3 and +4 meters.}
\label{fig:teaser}
\end{figure*}

The above mentioned problem can be mitigated by depth-based methods \cite{mlf, surfconv, d4lcn, pseudo, pseudo++, pseudo-e2e, patchnet, monofenet, am3d, monopsr}, where the depth information are utilized to provide more cues for bounding-box reasoning. For instance, the depth can be transformed into Pseudo-liDAR representations \cite{pseudo, pseudo++, pseudo-e2e, am3d, monofenet, monopl, patchnet}, fused with image features \cite{mlf}, or used to change the receptive field of convolutional kernels \cite{d4lcn} so that the detection model can capture more structural details of the objects. However, since the depth information is not available at the testing time, a depth estimator that trained from a collection of depth images should be plugged in the detection pipeline, making the depth-based methods hard to reach real-time speed.

In this work, we investigate a novel learning approach to improve the 3D object detection from depth images without training a cumbersome depth estimator. By utilizing the depth images, we augment the input data by synthesizing variety of images in the training stage. Unlike traditional data augmentation techniques that manipulate images in 2D space, we synthesize images at different virtual depths (Figure \ref{fig:teaser}), so that the detection model can learn more robust features to adapt to the depth changes of the same semantics. Therefore, there is no need to introduce a depth estimator in the testing stage. As depicted in Figure \ref{fig:concept}(b), the model trained with synthesized images can better estimate the object depth and achieve more accurate bounding-box localization.

To synthesize the images with virtual depth, we take advantage of novel-view synthesis methods \cite{synsin,view-control, extreme-view, view-stereo} and propose an efficient rendering module that uses depth image to reconstruct the 3D scene of the reference image, and synthesize images with virtual depth through the camera displacements. The ground-truth bounding-boxes are also shifted along the depth axis according to the displacement of current scene, producing synthetic training pairs for detection model to learn. There are several challenges for the detection model to learn from those synthetic images. First, the depth images are very sparse, making the synthetic images contain many holes but few meaningful semantics. Second, the provided depth images are not aligned well with the RGB images due to the calibration error, those unmatched depth pixels around the object boundaries make their appearance highly distorted and cause ghost artifacts in the synthetic image. To address these issues, we propose a multi-stage rendering module to generate high-quality synthesized images from coarse to fine.

In addition, we propose an auxiliary module in conjunction with the backbone of detection model. The auxiliary module consumes the backbone features for depth estimation and jointly optimize them with a pixel-wise loss. In this way, the detection model can learn more fine-grained details and achieve more reliable 3D reasoning. 

To the best of our knowledge, it is the first attempt to successfully employ synthetic images for 3D object detection. Our proposed learning framework will not introduce additional building blocks (e.g., depth estimator) and computation in the final deployment. The synthetic images can be readily learnt by other geometry-based detectors. Our contributions are summarized as follows:

\begin{itemize}

\item We propose a novel data augmentation strategy to ensure the effective learning of monocular 3D object detection by generating synthetic images with virtual depth.

\item We propose an efficient rendering module that can synthesize photo-realistic images from coarse to fine.

\item We propose an auxiliary module to jointly optimize the detection features through a pixel-level depth estimation task, leading to more accurate bounding-box reasoning.

\end{itemize}
We evaluate our learned 3D object detector on the KITTI \cite{kitti} 3D/BEV detection benchmark, and it demonstrates the state-of-the-art performance among the current monocular detectors.

\begin{figure*}[t]
   \centering
\begin{subfigure}[b]{0.66\textwidth}
\centering
 \includegraphics[height=5.0cm]{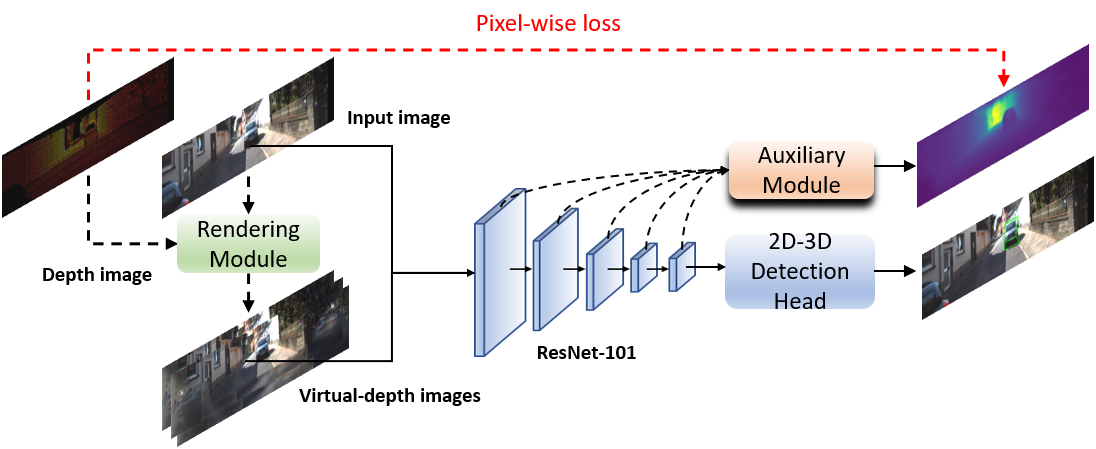}
 \caption{}
\end{subfigure}
~
\begin{subfigure}[b]{0.32\textwidth}
\centering
 \includegraphics[height=4.0cm]{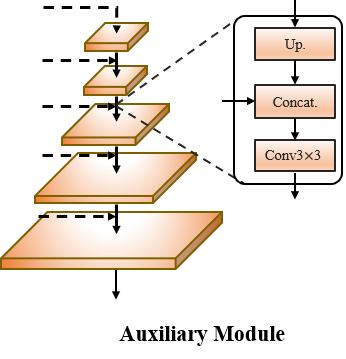}
\caption{}
\end{subfigure}

 \caption{(a) Overview of the proposed learning framework. The framework contains a detection model, a rendering module to synthesize virtual-depth images and a (b) pyramidal auxiliary module to transform the backbone feature for dense depth prediction.}
 \label{fig:model}
\end{figure*}

\section{Related work}
\noindent \textbf{Geometry-based monocular detection.}
The effective use of geometric priors can significantly improve the performance of monocular 3D object detection. Brazil \textit{et al.} \cite{m3d-rpn} proposed a 2D-3D anchoring mechanism, which employs 2D image features for 3D bounding-box detection. Mousavian \textit{et al.} \cite{geo3d} and Naiden \textit{et al.} \cite{shift-rcnn} utilized the 2D-3D perspective constraint to stabilize the predictions of 3D bounding-box. Liu \textit{et al.} \cite{smoke, rtm3d} proposed to estimate the 3D bounding-box in a group of key-points. Qin \textit{et al.} \cite{monogrnet} and Zhou \textit{et al.} \cite{monet3d} decoupled the detection task into multiple subtasks and optimized these subtasks in a global context. Chen \textit{et al.} \cite{monopair} utilized the spatial constrains between the object and its occluded neighbors, and proposed a pair-distance function to regularize the bounding-box prediction. Roddick \textit{et al.} \cite{oftnet} converted the 2D image features to bird-eye view using orthographic transformation, providing a holistically view of image features in 3D space.

In this paper, we demonstrate our proposed synthetic images can work well with three geometry-based detectors, SMOKE \cite{smoke}, M3D-RPN \cite{m3d-rpn} and RTM3D  \cite{rtm3d}.

\noindent \textbf{Depth-based monocular detection.} 
Depth images can effectively compensate for the lack of 3D cues in 2D images. Wang \textit{et al.} \cite{pseudo, pseudo++, monopl} explored to convert the depth image from monocular input into a pseudo-LiDAR representation, which can better capture the 3D structure of objects using existing point cloud detector. Qian \textit{et al.} \cite{pseudo-e2e} devised a soft-quantization technique to convert the depth image into the voxel input of point cloud detector, forming an end-to-end learning scheme. Xu \textit{et al.} \cite{patchnet} conducted an in-depth investigation of pseudo-LiDAR representation and proposed an efficient scheme based on 2D CNN with coordinate transformation. In additional to pseudo-LiDAR conversion, Xu \textit{et al.} \cite{mlf} introduced a multi-level fusion algorithm that progressively aggregated the features from RGB and depth images. Ding \textit{et al.} \cite{d4lcn} proposed a depth-aware model, where the convolutional receptive fields were dynamically adjusted by the depth image, distinctly processing the image contents at different depth levels. 

Unlike the above mentioned methods that require estimating depth information during the inference, our proposed framework only applies depth information in the training phase, thus presenting more efficient detection.

\noindent \textbf{Novel-view synthesis.} 
Novel-view synthesis has been widely studied in both vision and graphics. Early approaches \cite{infogan,inverse,view-single, view-flow} usually perform direct image-to-image transformation, enabling an end-to-end view manipulation. Later approaches \cite{synsin,view-control, extreme-view, view-stereo, interpretable, scene} focus on the learning of scene geometry, such as point cloud, multi-plane images, or implicit surfaces, from which continuous novel views with high degrees of freedom can be derived from the estimated 3D representations. 

In this work, considering that the monocular detectors are sensitive to the depth changes, we synthesize images by fixing the target view on the z-axis i.e., along depth direction, thereby avoiding the need to complement unknown information from extrapolated views.

\begin{figure*}[t]
\centering
 \includegraphics[width=\textwidth]{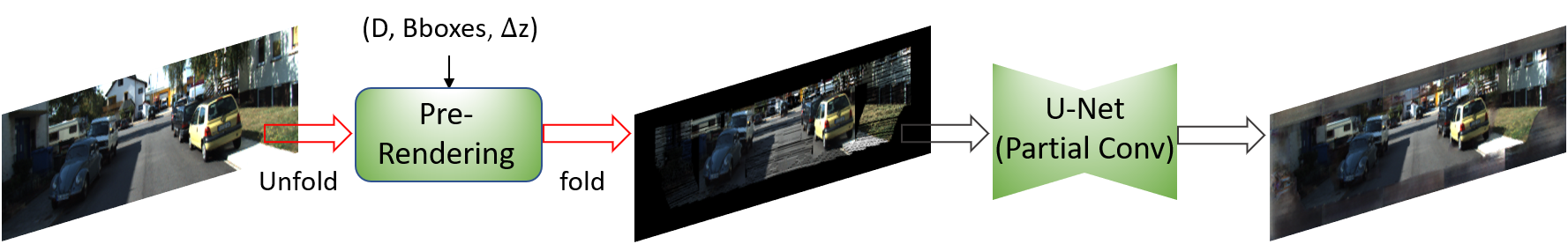}
 \caption{(a) The pipeline of rendering virtual-depth image. For background images, we first apply sparse rendering to generate virtual images with plausible semantics, and then a in-painting network to fill-in the missing areas. For foreground areas, we warp the 2D view-port of 3D boxes based on the perspective transformation from two camera poses.}
 \label{fig:render}
 \vspace{-2mm}
\end{figure*}

\begin{figure}[t]
\centering
 \includegraphics[width=\columnwidth]{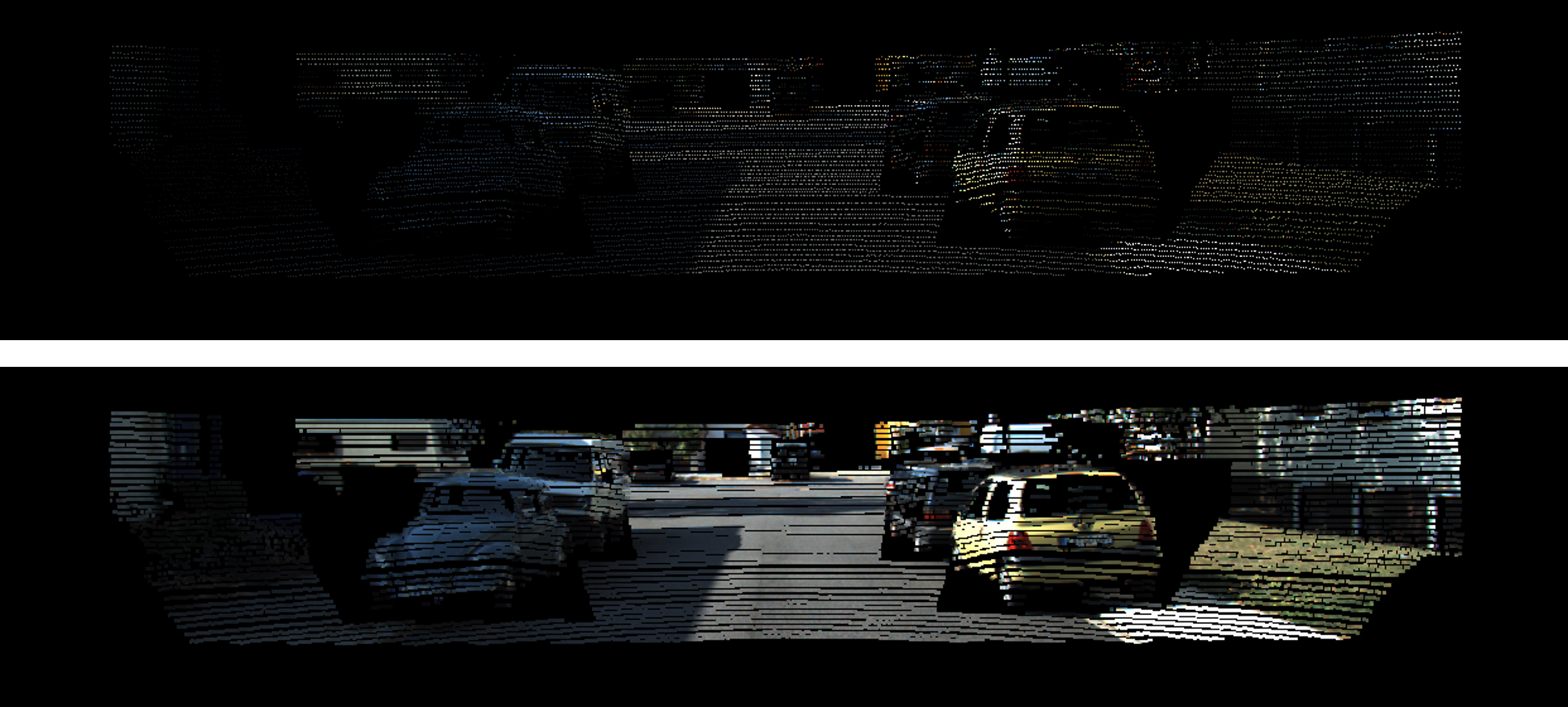}
 \caption{The naive renderings (top) and renderings from contextual image (bottom).}
 \label{fig:pre-render}
\end{figure}

\begin{figure}[t]
\centering
 \includegraphics[width=\columnwidth]{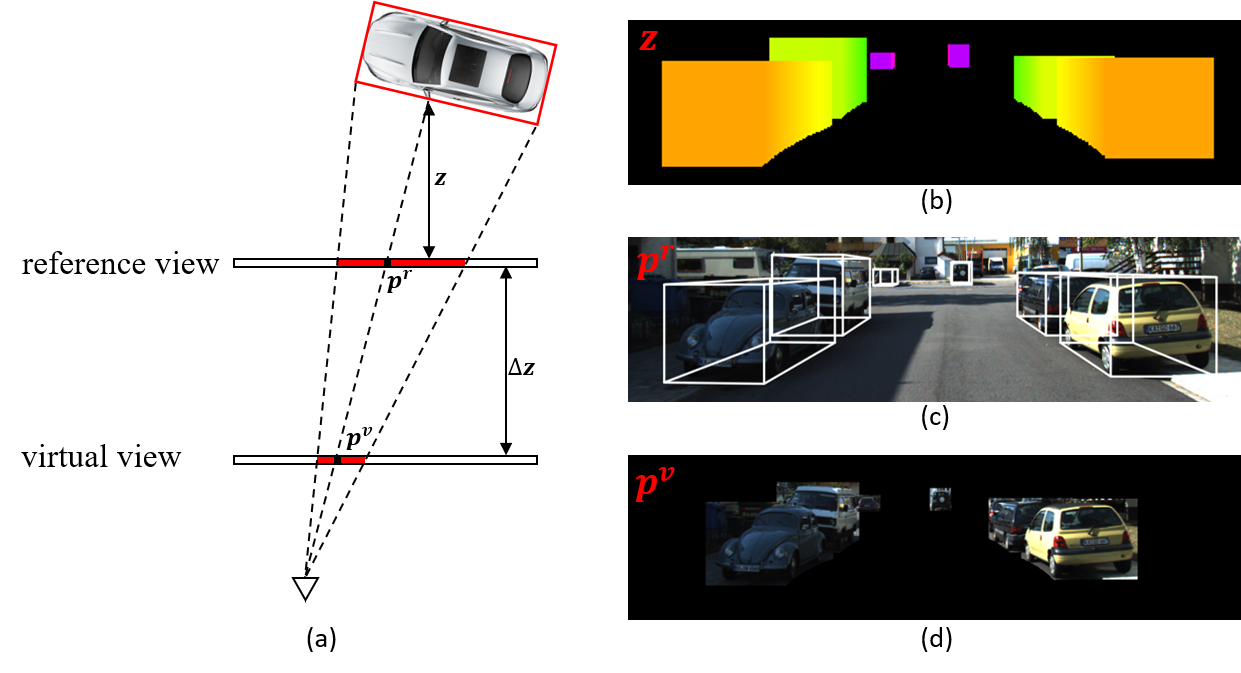}
 \caption{(a) Rendering foreground pixels on virtual image with bounding-box. (b) The foreground depth map, where the depth values are approximated by projecting the ground-truth 3D bounding-box (c) onto the image. (d) The foreground renderings on virtual image.}
 \label{fig:fg-render}
 \vspace{-2mm}
\end{figure}

\section{Methodology}
\subsection{Overall learning framework}
Our proposed learning framework is illustrated in Figure \ref{fig:model}. It is composed by three primary components: a detection model, a rendering module and an auxiliary module. On one hand, the rendering module takes as input the RGB image and its corresponding sparse depth image, produces a set of synthetic images at various virtual depth. All these synthetic images are used to train the detector as augmented data. On the other hand, the auxiliary module augments the backbone features of the detection model by jointly optimizing them with a pixel-wise loss, through a depth estimation task. Both rendering and auxiliary modules will be removed after training, and only the learned detector is deployed. In this paper, we employ an efficient detection model, termed Aug3D-RPN, which consist of a ResNet-101 \cite{resnet} backbone and a 2D-3D detection head \cite{d4lcn, m3d-rpn, kine3d}.


In the following subsections, we introduce the above components and the training scheme in detail.

\subsection{Synthesizing virtual-depth images for data augmentation}
\label{sec:render}
To improve the ability of monocular detector in discriminating object depth, we augment the training data by generating synthetic images with virtual depth. Here we use $\Delta z$ to specify the camera displacement between the reference and virtual views. The overall pipeline is shown in Figure \ref{fig:render}, in which we facilitate the depth information to perform virtual view synthesis and apply an in-painting network to fill-in the pixels with empty value. We first map the reference image into a 3D point cloud based on the nonzero values of its corresponding depth image $D$. For each position $(u,v)$ of the reference image, its 3D coordinate of point cloud can be calculated by:
\begin{equation}
    (x, y, z) = (\frac{u-c_x}{f}, \frac{v-c_y}{f}, 1) \cdot \mathds{1} [D_{(u,v)}>0],
    \label{eq:unproj}
\end{equation}
where $f$ and $(c_x, c_y)$ and $(f_x, f_y)$ denote the focal length and the principle points of the reference camera, respectively. The virtual-depth image can then be synthesized by projecting the point cloud from the current camera coordinates to the image plane in virtual view. Given the camera projection matrix $\textbf{P}$, the position $(u',v')$ of each point after projection can be calculated by:
\begin{equation}
    (z+\Delta z) \cdot [u', v', 1]^T = \textbf{P} \cdot [x, y, z+\Delta z, 1]^T
    \label{eq:proj}
\end{equation}
The rendering process can then be regarded as transporting each colored pixel from $(u, v)$ to $(u',v')$.

However, as the depth image is extremely sparse, the output image through the above sparse rendering method is dominated by tremendous black pixels. The limited semantics in the resulted renderings make the in-painting network hard to train. In addition, some depth images are not aligned well with the reference images due to the inevitable calibration error. The unmatched pixels will distort the object's appearance and yield ghost artifacts at the object boundaries.

To address the above issue, we perform a pre-rendering step to enhance the density of renderings. Specifically, we ``unfold'' the reference image into a contextual image by stacking its $3\times 3$ neighboring pixels of each position into a single channel. Given an image of shape $H\times W\times 3$, the contextual image has a shape of $H\times W\times 27$. Then we applied the above rendering formula to extract the sparse version of contextual image and then ``fold'' the output back to the RGB space. As displayed in Figure \ref{fig:pre-render}, the renderings from contextual image is more dense and full of meaningful contents. 

As for foreground objects, since they mainly have rigid body, we can use the surface of the ground-truth bounding-box to approximate their depth. As shown in Figure \ref{fig:fg-render}, we first calculate the depth values in the 2D ``viewport'' of these objects based on the geometries of their bounding-boxes. Then,  based on these depth values, we can calculate the appearance flow for each foreground pixel according to Eq. \ref{eq:unproj} and Eq.\ref{eq:proj}, and remap them on the synthetic image.

After that, we employ a standard U-Net architecture with partial convolutions \cite{partial-conv} to complete the color pixels from missing areas. To train the in-painting network, we let the detection images as training targets, and their pre-renderings with $\Delta z=0$ as training inputs. We use pixel-wise $\ell_1$, perceptual loss and adversarial loss $\mathcal L_{adv}$ to supervise the in-painting network. Given a reference image $I_{ref}$ and its pre-renderings $I_{pre}$, we have:
\begin{equation}
    \mathcal L = ||I_{ref}-I_{pre}||_1 + ||\phi(I_{ref})-\phi(I_{pre})||_2^2 + \mathcal L_{adv}(I_{ref}, I_{pre}),
\end{equation}
where $\phi$ is the hidden feature activation from a pre-trained VGG-16 classification network. $\mathcal L_{adv}$ is a relativistic adversarial loss from \cite{inpaint}, in which a global and a local discriminator are utilized for perception enhancement. We leave the configuration of our in-painting network and more training details to the \textbf{supplementary material}.

\subsection{Joint depth estimation for feature augmentation.}
\label{sec:aux}
One of the core idea of this work is utilizing pixel-level supervision to improve the understanding of object-level task, which has been proven effective in many prior arts \cite{sassd,mmf,rock}. Based on this, we propose a detachable auxiliary module to augment the backbone features by jointly optimizing depth estimation and object detection tasks. In particular, we use a pyramidal architecture to deploy our auxiliary module, so that it can aggregate multi-scale detection features into a full-resolution representation. For ResNet-101, we fuse the outputs from the first convolutional layer \textit{Conv1} and four residual blocks \textit{layer1}, \textit{layer2}, \textit{layer3}, \textit{layer4}. 

As depicted in Figure \ref{fig:model} (b), we upsample the feature from the bottom stage with bilinear interpolation. The features across stages are first concatenated and then undergone a $3\times3$ convolutional layer to generate feature with 256 channels. In the final stage, one additional 1×1 convolution is applied to produce depth estimation $\tilde D$. Since the foreground components and background components have different depth distributions, we applied the following weighted loss to jointly optimize the detection and depth estimation tasks:
\begin{equation}
    \mathcal L = \mathcal L_{\text{det}} +  \lambda_{fg} \frac{1}{N_i} \mathcal L(\tilde D_i, D_i) + \lambda_{bg} \frac{1}{N_j} \mathcal L( \tilde D_j, D_j)
    \label{eq:depth}
\end{equation}
where $\mathcal L_{\text{det}}$ is the detection loss in \cite{m3d-rpn}, $D$ is the ground-truth depth image, $i$ and $j$ respectively represent the nonzero foreground and background pixels in the ground-truth depth image. We found that by setting $\lambda_{fg}=2.5$, $\lambda_{bg}=1$ and $\mathcal L$ with $\text{Smooth-}l_1$ \cite{ssd} loss, our detection model can generally achieve the best performance. 

\subsection{3D object reasoning with 2D-3D detection head}
The detection head predicts both 2D and 3D geometries of the bounding-boxes based on a set of pre-defined 2D-3D anchors, and then associates these geometries according to the perspective relationship. 

Specifically, we define a 2D-3D anchor by $(a^{w}_{2D}, a^{h}_{2D})\text{-}(a^{w}_{3D}, a^{h}_{3D}, a^{l}_{3D}, a^{z}, a^\theta)$, where $(a^{w}_{2D}, a^{h}_{2D})$ denote its width and height in image scale, $(a^{w}_{3D}, a^{h}_{3D}, a^{l}_{3D})$, $a^{z}$ and $a^{\theta}$ represent its 3D dimensions, depth and observation angle in camera coordinates, respectively. 
Given 2D predictions $(p^{u}_{2D}$, $p^{v}_{2D}$, $p^{w}_{2D}$, $p^{h}_{2D})$, the 2D bounding-box $(b^x_{2D}, b^y_{2D}, b^w_{2D}, b^h_{2D})$ at each anchor position $(u,v)$, can be written as:
\begin{align}
& b^x_{2D} = u+a^{w}_{2D}p^{u}_{2D}, \quad b^y_{2D}=v+a^{h}_{2D}p^{v}_{2D},\\
& b^{w}_{2D} = e^{a^{w}_{2D}p^{w}_{2D}},\quad b^{h}_{2D} = e^{a^{h}_{2D}p^{h}_{2D}}. 
\end{align}
Given 3D predictions $(p^{w}_{3D}$, $p^{h}_{3D}$, $p^{l}_{3D}$, $p^{\Delta z}$, $p^{\theta})$, the dimension of 3D bounding box $(b^h_{3D}, b^w_{3D}, b^l_{3D}, b^{\theta})$, in terms of height, width, length and observation angle, can be given as follows:
\begin{align}
    &b^w = e^{a^{w3d}p^{w3d}}, \quad b^h = e^{a^{h3d}p^{h3d}}, \\
    &b^l = e^{a^{l3d}p^{l3d}}, \quad b^{\theta} = \Omega(a^{\theta}, p^{\theta} ),
\end{align}
Its location $(b^x_{3D}, b^y_{3D}, b^z_{3D})$ in camera coordinates therefore subjects to
\begin{equation}
    (a^{z} + p^{\Delta z})  \cdot [b^x_{2D}, b^y_{2D}, 1]^T = \textbf{P} \cdot [b^x_{3D}, b^y_{3D}, b^z_{3D}, 1]^T \\
\end{equation}
where $\textbf{P}\in {\rm I\!R}^{3\times 4}$ is the camera projection and $\Omega(\cdot)$ serves to round the radian value within $[-\pi, \pi]$. 

\section{Experiments}

\begin{table*}[t]
\caption{Performance comparison with state-of-the-art methods on KITTI test set. BEV and 3D object detection metric are used, reported by AP40 (IoU$>$0.7). The top three performers are indicated by \textcolor{red}{red}, \textcolor{blue}{blue} and \textcolor{green}{green} colors, respectively. The runtime is reported from the KITTI leaderboard with slight variances in hardware.}
\centering
\begin{adjustbox}{width=0.90\textwidth}
\begin{tabular}{c|c|c|ccc|ccc|c}
\hline
\multirow{2}{*}{Method} &
\multirow{2}{*}{Conference} &
\multirow{2}{*}{\#Frames} &
\multicolumn{3}{c|}{\textit{3D}} & \multicolumn{3}{c|}{\textit{BEV}} &
\multirow{2}{*}{runtime (s/img)} \\ 
\cline{4-9}
{} & {} & {} & Easy & Moderate  & Hard  & Easy & Moderate & Hard  & {}  \\ \hline\hline

FQNet\cite{fqnet} &CVPR19 &1 &2.77 & 1.51 & 1.01 & 	5.40 & 3.23 & 	2.46 & 0.5\\

ROI-10D\cite{roi10d} &CVPR19 &1 & 4.32 &2.02& 1.46& 	9.78 & 	4.91 &3.74  & 0.2\\

GS3D\cite{gs3d} &CVPR19 &1 & 4.47& 2.90 & 2.47& 	8.41 & 6.08 &	4.94  & 2.3\\

Shift R-CNN \cite{shift-rcnn} &ICIP19 &1 &6.88&3.87&2.83&11.84 & 6.82&5.27 & 0.25\\

MonoFENet\cite{monofenet} &TIP19 &1 & 8.35 &5.14 & 	4.10 & 	17.03 &11.03 &	9.05  & 0.15\\

MLF \cite{mlf} &CVPR18 &1 &7.08 &5.18 &4.68 &-&-&-&0.12\\

MonoGRNet\cite{monogrnet} &AAAI19 &1 & 9.61 & 5.74 & 4.25 & 18.19 & 11.17 & 8.73 & 0.04\\

MonoPSR \cite{monopsr} & CVPR19 &1 & 10.76 & 7.25 & 5.85 & 18.33 & 12.58 &	9.91 & 0.2 \\

MonoPL \cite{monopl} & CVPR19 &1 &10.76 & 7.50  & 6.10 & 21.27 &	13.92 &	11.25& - \\

MonoDIS \cite{monodis} & ICCV19 &1 &10.37 & 7.94 &6.40 & 17.23 & 13.19 & 11.12 & - \\

M3D-RPN\cite{m3d-rpn} & ICCV19 &1 &14.76 & 9.71 &	7.42 &	21.02 &13.67 &10.23 & 0.16\\

SMOKE \cite{smoke} & CVPRW20 &1 &14.03 & 9.76 & 	7.84 & 	20.83&14.49 &	12.75& 0.03 \\

MonoPair \cite{monopair} &CVPR20 &1 & 13.04 & 9.99 & 	8.65 &  19.28 &	14.83 & 12.89 & 0.06 \\

AM3D \cite{am3d} &ICCV19 &1 &	16.50 & 10.74 &	\textcolor{green}{9.52} & 	\textcolor{green}{25.03} & \textcolor{green}{17.32} & \textcolor{blue}{14.91} & 0.4\\	

RTM3D \cite{rtm3d} &ECCV20 &1 & 14.41 & 10.34 &	8.77 & 	19.17 & 14.20 & 	11.99 & 0.05 \\

PatchNet \cite{patchnet}&ECCV20 &1 &15.68 & 11.12 & \textcolor{red}{10.17} & 22.97 & 16.86 & \textcolor{red}{14.97} & 0.4\\

D4LCN \cite{d4lcn} &CVPR20	&1 & \textcolor{green}{16.65} & \textcolor{green}{11.72} & 	9.51  & 	22.51 & 16.02 &	12.55 & 0.6\footnotemark[2]\\

Kinematic3D \cite{kine3d} &ECCV20 &4	& \textcolor{red}{19.07} & \textcolor{blue}{12.72} & 	9.17& \textcolor{red}{26.69} &	\textcolor{blue}{17.52} & 	13.10 & 0.12 \\

\hline \hline
Aug3D-RPN (ours) &  -& 1 &	\textcolor{blue}{17.82} & 	\textcolor{red}{12.99} & 	\textcolor{blue}{9.78} & \textcolor{blue}{26.00} & 	\textcolor{red}{17.89} & 	\textcolor{green}{14.18} & 0.08\\\hline
\end{tabular}
\end{adjustbox}
\label{tbl:overall}
\vspace{-3mm}
\end{table*}

We use KITTI dataset \cite{kitti} to evaluate our proposed Aug3D-RPN. The dataset contains 7,481 training samples and 7,518 testing samples. To evaluate the effectiveness of different modules, we use the splits in \cite{mv3d} by dividing the training set into 3,712 training samples and 3,769 validation samples.

There are two detection tasks, known as 3D detection (\textit{3D}) and birds-eye-view localization (\textit{BEV}) tasks. Each task has three levels of difficulties: \textit{easy}, \textit{moderate} and \textit{hard}, according to the object size, occlusion and truncation level, and the ranking is based the average precision (AP) on the \textit{moderate} level. The AP with 11 points \cite{3dop} and 40 points \cite{monodis} interpolation are respectively denoted by AP11 and AP40.

\subsection{Implementation details} \label{sec:settings}
We first extract the depth images following \cite{pseudo++, pseudo-e2e}. Both depth images and RGB images are padded to $384\times 1248$ and their mirrored versions are randomly sampled with a probability of 0.25 during training. For each ground-truth 3D box, we calculate its 2D box by projecting it onto the image plane. We ignore those objects which have 2D boxes less than 16 pixels or larger than 256 pixels in height, and those with visibility less than 0.5. 

To build the anchor classification target, we calculate the IoU between the 2D anchors and 2D ground-truths. An IoU threshold of 0.5 is used to label the anchors as positive or negative. The anchors that have IoU greater than 0.5 with the ignore region, will be ignored. Besides, we apply online hard negative mining (OHNM) \cite{ssd} to re-balance the positive and negative anchor targets with a ratio of 1:3. 

To create virtual-depth images, we ignore those objects that are truncated by camera or have visibility less than 0.6. By removing the depth values of these objects, our in-painting network will complete their area as background. For each training iteration we sample 2 camera displacements from a uniform distribution $U(0, 5)$ and 1 camera displacement from $U(-1, 0)$. The later is to increase the number of objects that truncated by camera.

Our Aug3D-RPN detector is trained for 60 epochs by using the SGD optimizer. The batch size, learning rate, and weight decay are set to 4, 0.004, and 0.0005, respectively. The learning rate is decayed with a cosine annealing strategy \cite{sgdr}. In the inference phase, we select the predicted boxes with confidence larger than 0.75 and apply non-maximum suppression (NMS) with an IoU threshold of 0.4 to remove the duplicated boxes.

\begin{table*}[t]
\caption{Evaluation of 3D/BEV detection performance on KITTI val set, reported by AP11. The abbreviation ``Syn.'' indicates using synthetic images for training. ``$\ast$'' indicates the results are reproduced by the official public codes, which are slightly different from the results in the paper.}
\centering
\begin{adjustbox}{width=0.99\textwidth}
\begin{tabular}{c|ccc|ccc|ccc|ccc}
\hline
\multirow{2}{*}{Method} &

\multicolumn{3}{c|}{\textit{3D} (IoU$>$0.7)} & \multicolumn{3}{c|}{\textit{BEV} (IoU$>$0.7)} &
\multicolumn{3}{c|}{\textit{3D} (IoU$>$0.5)} & \multicolumn{3}{c}{\textit{BEV} (IoU$>$0.5)} \\
\cline{2-13}
{} & Easy & Moderate  & Hard  & Easy & Moderate & Hard & Easy & Moderate  & Hard  & Easy & Moderate & Hard  \\ \hline\hline
$\text{M3D-RPN} ^{\ast}$\cite{m3d-rpn} & 20.88 & 17.39 & 15.51 & 27.13 & 21.71 & 18.30 & 50.17 & 40.27 & 33.62 & 58.21 & 43.68 & 36.23 \\

M3D-RPN + Syn.          & 24.41 & 20.52 & 17.13 & 31.86 & 26.12 & 20.12 & 54.56 & 43.72 & 35.25 & 63.68 & 47.99 & 36.79 \\ \hdashline
Delta                 & 3.53  & 3.13  & 1.62  & 4.73  & 4.41  & 1.82  & 4.39  & 3.45  & 1.63  & 5.47  & 4.31 & 0.56 \\ \hline

$\text{SMOKE} ^{\ast}$\cite{smoke}    & 18.32 & 15.84 & 15.23 & 24.39 & 20.64 & 17.37    & 50.08 & 37.39 & 32.87 & 53.93 & 39.98 & 39.20 \\

SMOKE + Syn. & 21.46 & 18.31 & 16.60 & 28.11 & 24.12 & 18.99    & 54.23 & 41.02 & 35.22 & 56.98 & 44.55 & 40.47 \\ \hdashline

Delta    & 3.14  & 2.47  & 1.37  & 3.72  & 3.48  & 1.62     & 4.15  & 3.63  & 2.35  & 3.05  & 4.57  & 1.27 \\ \hline

RTM3D\cite{rtm3d} & 19.19 & 16.70 & 16.14 & 25.96 & 21.88 & 18.88    & 54.97 & 42.68 & 36.95 & 60.98 & 45.74 & 42.93\\

RTM3D + Syn. & 22.12 & 19.29 & 17.25 & 30.21 & 25.37 & 20.28    & 58.72 & 45.77 & 38.27 & 64.62 & 49.91 & 44.15 \\ \hdashline

Delta    & 2.93  & 2.59  & 1.11  & 4.25  & 3.49  & 1.40     & 3.75  & 3.09  & 1.32  & 3.64  & 4.17  & 1.22 \\
\hline
\end{tabular}

\end{adjustbox}

\label{tbl:delta}
\vspace{-3mm}
\end{table*}

\subsection{Comparison with the state-of-the-art}
We compare our Aug3D-RPN detector with other state-of-the-art detectors by submitting the detection results to the KITTI server for evaluation. The evaluation is based on AP40 and the results are presented in Table \ref{tbl:overall}. As can be seen, our method ranks higher than all previous single-frame methods in both 3D and BEV metric. The performance on easy level are comparable to the best performers Kinematic3D \cite{kine3d}, which is a multi-frame detector.  Besides, our method only runs at 80ms per image, which is more than five times faster than the recently proposed detectors, such as AM3D ($\sim$400ms), D4LCN ($\sim$600ms) and PatchNet ($\sim$400ms), as all these competitors are depth-based detectors and they all rely on an expensive dense depth input. In contrast, Aug3D-RPN is trained to be depth-aware without explicitly requiring depth information in the testing stage. The ResNet backbone and detection head take only around 50ms and 30ms in the inference, respectively. \footnotetext[2]{The runtime reported on the KITTI leaderboard does not count the runtime of depth estimation, we approximate its overall runtime based on its official code.}

Comparing with those geometry-based monocular 3D detectors, such as SMOKE\cite{smoke}, M3D-RPN\cite{m3d-rpn}, and RTM3D\cite{rtm3d}, Aug3D-RPN presents significant advantages. Aug3D-RPN outperforms these methods by up to $(3.1\%, 2.7\%, 1.0\% )$ in 3D detection and $(5.0\%, 3.7\%, 2.2\% )$ in BEV detection. The performance on KITTI val set are also shown in Table \ref{tbl:val}. Again, our method can achieve state-of-the-art performance on most detection metrics, which demonstrates the effective learning with synthetic images and auxiliary module.

\begin{table}[t]
\caption{3D object detection performance on KITTI val set, reported by AP11. ``-'' means that the results are not available. The top 2 performers are indicated by \textcolor{red}{red} and \textcolor{blue}{blue} colors, respectively.}
\centering
\begin{adjustbox}{width=1\columnwidth}
\begin{tabular}{c|ccc}
\hline
\multirow{2}{*}{Method} &
\multicolumn{3}{c}{\textit{3D / BEV} (IoU$>$0.7)} \\ 

\cline{2-4}
{} & E & M & H  \\ \hline\hline
MonoPSR \cite{monopsr} &12.75 / 20.63 & 11.48 / 18.67 &8.59 / 14.45 \\
MonoDIS \cite{monodis} &18.05 / 24.26 & 14.98 / 18.43 &13.42 / 16.95 \\
SMOKE \cite{smoke}     &14.76 / 19.99 & 12.85 / 15.61 &11.50 / 15.28 \\
M3D-RPN \cite{m3d-rpn} &20.27 / 25.94 & 17.06 / 21.18 & 15.21 / 17.90 \\
RTM3D \cite{rtm3d}     &19.19 / 25.56 & 16.70 / 22.12 &16.14 / 20.91 \\
AM3D\cite{am3d}    &\textcolor{red}{32.23} / - & 21.09 / - & 17.26 / -   \\
D4LCN \cite{d4lcn} & 26.97 / \textcolor{blue}{34.82} & \textcolor{blue}{21.71} / \textcolor{blue}{25.83} & \textcolor{blue}{18.22} / \textcolor{blue}{23.53} \\

\hline\hline
Aug3D-RPN (ours) & \textcolor{blue}{28.53} / \textcolor{red}{36.27}  & \textcolor{red}{22.61} / \textcolor{red}{28.76} & \textcolor{red}{18.34} / \textcolor{red}{23.70} \\\hline
\end{tabular}
\end{adjustbox}
\label{tbl:val}
\vspace{-2mm}
\end{table}

\begin{table}[t]
\centering
\caption{Ablation Experiments on the KITTI val set using AP11. ``Syn'' means using synthetic images for training, ``Aux-fg'' and ``Aux-bg'' refer to imposing auxiliary loss on the foreground and background image areas, respectively.}
\begin{adjustbox}{width=0.95\columnwidth}
\begin{tabular}{c|ccc|ccc}
\hline
\multirow{2}{*}{Exp.} &
\multirow{2}{*}{Syn.} &
\multirow{2}{*}{Aux-fg.} &
\multirow{2}{*}{Aux-bg.} &
\multicolumn{3}{c}{\textit{3D} (IoU $>$0.7)} \\ 
\cline{5-7}
 {} & {} & {} & {}  & E & M & H  \\ \hline\hline
1 & &  &  & 22.62 & 17.44 & 15.48 \\
2 & \checkmark &  &  & 25.89 & 20.45 & 17.22  \\
3 & & \checkmark & & 24.51& 18.91& 16.63 \\
4 & & & \checkmark &23.14 &17.98 &16.25 \\
5 & & \checkmark & \checkmark &25.16 &19.48& 17.31 \\
6 & \checkmark & \checkmark & \checkmark &28.53 & 22.61 & 18.34 \\
\hline
\end{tabular}
\end{adjustbox}
\label{tbl:abl}
\vspace{-2mm}
\end{table}

\begin{figure}[t!]
\centering
\begin{subfigure}{0.49\columnwidth}
 \includegraphics[width=\columnwidth]{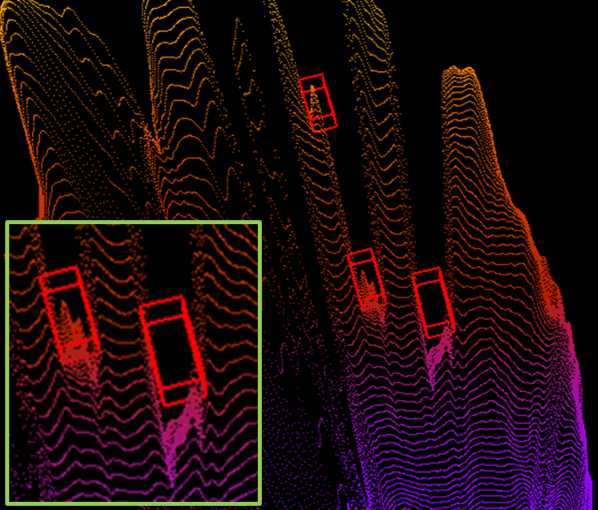}
\caption{}
\end{subfigure}
\begin{subfigure}{0.49\columnwidth}
 \includegraphics[width=\columnwidth]{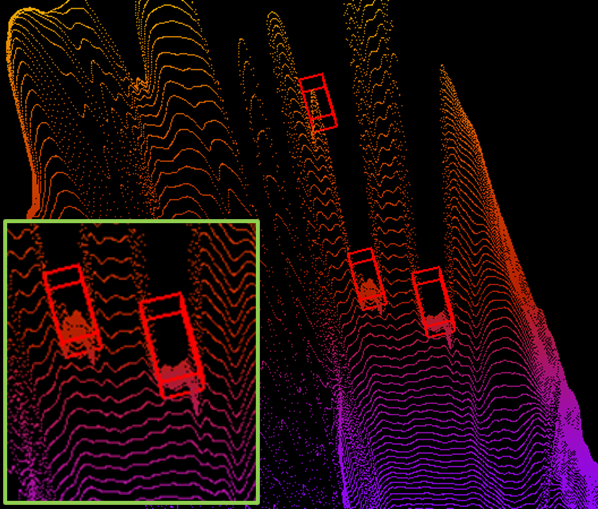}
\caption{}
\end{subfigure}
\caption{The Pseudo-LiDAR generated by projecting the depth estimation from (a) Exp.5 and (b) Exp.6. The ground-truth bounding-boxes are shown in red.}
\label{fig:plidar}
\end{figure}

\subsection{Ablation study}
\label{sec:ablation}
We first demonstrate our synthetic virtual-depth images can work well with other geometry-based detectors. Here we choose three baseline detectors, M3D-RPN \cite{m3d-rpn}, SMOKE \cite{smoke} and RTM3D \cite{rtm3d}. All of them can achieve real-time efficiency and with neat architecture. By running their public codes, we present their performances on KITTI val set. As shown in Table \ref{tbl:delta}, the model trained with virtual-depth images can consistently achieve around 2$\sim$3 points (in AP11) improvements on easy and moderate levels. The slight improvement on hard level is because most objects in this category have very few depth values due to high occlusion and long distance, which makes them hard to be rendered in the synthetic images.

Then we perform a in-depth analysis of how different components contribute to Aug3D-RPN.
\begin{figure*}[t]
    \centering
    \includegraphics[width=0.99\textwidth]{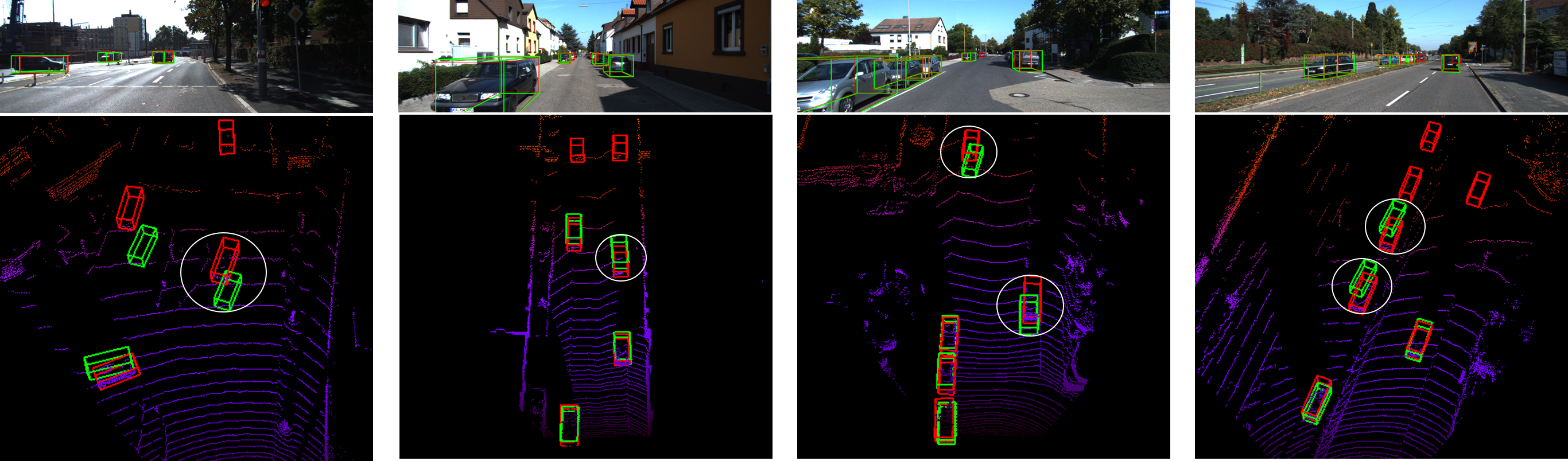}
      \vspace{3mm}
    (a) The predictions by the baseline detection model. 
    \includegraphics[width=0.99\textwidth]{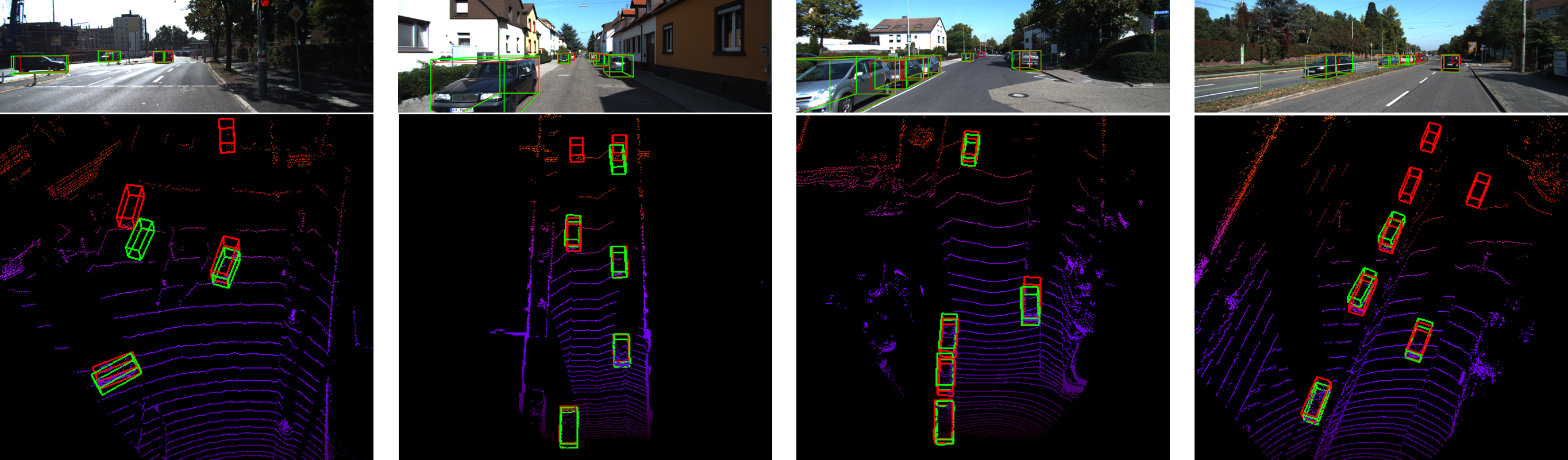}
     \vspace{3mm}
    (b) The predictions by the proposed Aug3D-RPN model.
    \caption{Examples of detection results on KITTI val set. The predicted and ground-truth bounding boxes are shown in green and red colors, respectively. The predictions on 3D point cloud are plotted for better visualization. Best viewed in color.}
    \label{fig:qualitative}
    \vspace{-6mm}
\end{figure*}

\begin{table}
\centering
\begin{adjustbox}{width=0.95\columnwidth}
\begin{tabular}{c|cccc}
\hline
   Method          & 0-10m  & 10-20m & 20-30m  & 30-40m \\  \hline
   w/o Syn.           & 2.05 & 3.29  & 5.07 & 9.41 \\   
   w/ Syn.            & 1.47 & 2.18  & 4.23 & 8.80  \\
Relative Delta.       & -28.3\% & -33.7\% & -16.6\% & -6.5\%  \\
\hline 
\end{tabular}
\end{adjustbox}
\caption{The absolute depth error (in meters) in different depth intervals.}
\label{tbl:depth}
 \vspace{-5mm}
\end{table}

\noindent \textbf{Effectiveness of synthetic virtual-depth images.} 
As presented in Table \ref{tbl:abl}, the model learning with synthetic images can achieve $(3.3\%, 3.0\%, 1.7\%)$ improvements by the baseline model (Exp.2 vs. Exp.1), and $(3.4\%, 3.2\%, 1.0\%)$ by the auxiliary-driven model (Exp.6 vs. Exp.5). On top of the auxiliary-driven model, we also evaluate how virtual-depth images can boost the overall depth estimation in pixel-level. We calculate the absolute depth error (absError) within a reliable depth range (0m, 40m), the error rates (in meters) of each 10m interval are reported in Table \ref{tbl:depth}. As can be seen, the model trained with synthetic images can exhibit more accurate depth estimation. Fig. \ref{fig:plidar} presents a qualitative comparison by projecting the estimation into 3D space, in the form of Pseudo-LiDAR. As can be seen, the model trained with synthetic data can obtain more compact 3D points attached to the bounding-box boundary, which means that the depth prediction of the object is more accurate and consistent. 

\noindent \textbf{Effectiveness of auxiliary module.} 
As shown in Table \ref{tbl:abl}, the model guided by auxiliary module can earn additional $(2.6\%, 2.2\%, 1.1\%)$ points (Exp.6 vs. Exp.2). Actually, we can apply depth estimation loss separately on foreground/background regions of the input image. As shown in Table \ref{tbl:abl}, applying loss on the foreground regions can result in $(1.9\%, 1.5\%, 1.2\%)$ performance gains, while applying loss on background regions can improve the performance by $(0.5\%, 0.5\%, 0.8\%)$. This reveals the fact that both images components are important to support the reasoning of object depth. By discriminating the depth of foreground object, the model can be aware of its appearance, therefore better inferring its orientation and local geometry. The background of images encode the geometry of the scene, which provides additional contextual cues to estimate the object depth. 

\noindent \textbf{Qualitative results.} 
In Figure \ref{fig:qualitative}, we demonstrate some predictions from the baseline and full-setting models (Exp.1 vs. Exp.6). As can be observed, the detector resulted from our learning framework can predict more reliable 3D bounding boxes without having additional memory and runtime.

\section{Conclusion}
In this paper, we propose a effective learning framework, which consist of a rendering module and an auxiliary module, to improve the monocular 3D object detection. The rendering module augments the training data by synthesizing images with virtual depths, from which the detector can learn to discriminate the depth changes of the objects. The auxiliary module can guide the detection model to learn better structure information about objects and scenes. Both modules significantly improve the detection accuracy and then are removed in the final deployment, adding no computational cost. Experiments on the KITTI 3D/BEV detection benchmark demonstrate the synthetic images can work well with existing monocular detectors. The proposed Aug3D-RPN can achieve leading accuracy and more significantly be $>$4 times faster than the recent state-of-the-art methods.

{\small
\bibliographystyle{ieee}
\bibliography{main}
}

\end{document}